\pdfoutput=1

\documentclass[11pt]{article}

\usepackage[]{EMNLP2023}

\usepackage{times}
\usepackage{latexsym}

\usepackage[T1]{fontenc}

\usepackage[utf8]{inputenc}

\usepackage{microtype}

\usepackage{inconsolata}

\usepackage{graphicx}
\usepackage{cleveref}
\usepackage{enumitem}
\usepackage{subcaption}
\usepackage{booktabs}

\newcommand\blfootnote[1]{%
  \begingroup
  \renewcommand\thefootnote{}\footnote{#1}%
  \addtocounter{footnote}{-1}%
  \endgroup
}

\title{MidiTok: A Python package for MIDI file tokenization}

\author{Nathan Fradet$^{1,2}$, Jean-Pierre Briot$^{1}$, Fabien Chhel$^{4,3}$,\\
        \textbf{Amal El Fallah Seghrouchni}$^{1}$\textbf{, Nicolas Gutowski}$^{3}$ \\
        $^{1}$Sorbonne University, CNRS, LIP6, F-75005 Paris \\ $^{2}$Aubay, Boulogne-Billancourt, France \\ $^{3}$University of Angers, LERIA, 49000 Angers, France \\ $^{4}$ESEO, ERIS, 49100 Angers, France }

\begin{document}
\maketitle
\begin{abstract}
Recent progress in natural language processing has been adapted to the symbolic music modality. Language models, such as Transformers, have been used with symbolic music for a variety of tasks among which music generation, modeling or transcription, with state-of-the-art performances. These models are beginning to be used in production products. To encode and decode music for the backbone model, they need to rely on tokenizers, whose role is to serialize music into sequences of distinct elements called tokens.
MidiTok is an open-source library allowing to tokenize symbolic music with great flexibility and extended features. It features the most popular music tokenizations, under a unified API. It is made to be easily used and extensible for everyone. \url{https://github.com/Natooz/MidiTok}
\end{abstract}

\blfootnote{MidiTok was initially presented at the ISMIR 2021 late breaking demos \cite{miditok2021}. This document is an updated and comprehensive report.}

\section{Introduction}\label{sec:miditok_introduction}

Recent progress in natural language processing (NLP), such as Transformers \cite{attention_is_all_you_need}, has been used with symbolic music for several tasks such as generation \cite{musictransformer2018,huang_remi_2020,vonrutte2022figaro}, understanding \cite{zeng2021musicbert}, or transcription \cite{hawthorne2021sequencetosequence,mt3}, with state-of-the-art performances. Most generative deep learning models for symbolic music are nowadays based on LMs. Using LMs for symbolic music requires, as for natural language, to tokenize the music, i.e. serialize it into sequences of distinct tokens. These tokens will represent note attributes such as pitch or duration, and time events.

The tokenization of music, i.e. the conversion of notes into sequences of tokens, is however not a straightforward process. Unlike text, polyphonic music comes with simultaneous notes, each of them having several properties, and the problem becomes even more complex if we consider several tracks or instruments.
Many recent research papers introduced different ways to tokenize symbolic music, but few authors share they source code in an easy way to reproduce, or to just use their methods. Furthermore, music files, such as MIDIs, need to be properly preprocessed. This step consist in downsampling its values, such as the onset and offset times of notes, their velocities, duration, or tempo values among others.

With the motivation to offer a friendly and convenient way to tokenize MIDI files, we created MidiTok. It implements the most popular tokenizations, under a unified API. It offers a great flexibility and extended features, so one can easily train and use LMs for symbolic music, and compare the different tokenizations. MidiTok was first introduced in late 2021 \cite{miditok2021}, and in the meantime received multiple updates until it became established in the community. Today, it is the "go-to" solution for researchers and engineers to use LMs with symbolic music. MidiTok has been built all along the making of this thesis, and has been used for most of the results reported in it. MidiTok is a major living contribution of this thesis, that we believe will continue to evolve.

In the next section, we describe the overall workflow of MidiTok, then the tokenizations and features it implements, and finally some user insights.

\section{Tokenizing music}\label{sec:miditok_tokenizing_music}

In this section, we introduce the data contained in MIDI files, and the previous works which tokenized symbolic music.

\subsection{The data in MIDI files}

When thinking about tokenizing symbolic music, we first need to think about what information present in the MIDI files should be considered.
A MIDI file contains several categories of contents. The most important are: tracks of instruments, tempo changes, time signature changes, key signature changes and lyrics. Tracks themselves contain notes - more precisely their pitch, velocity, onset and offset times - and effects such as sustain pedal, pitch bend and control changes. You can find the complete MIDI specifications on the MIDI Manufacturers Association website\footnote{\url{https://www.midi.org/specifications}}.

All of these information come as the form of events that occur a certain moments in time. The base time unit of MIDIs is the \textit{tick}, which resolution is called the time division and expressed in ticks per quarter note (or ticks per beat when the time signature is $\frac{*}{4}$). The time division is usually a multiple of 12 and a high value. Common values are 384 and 480. Hence, these events happen each at certain \textit{tick}, and we will have to express them as tokens, along with \textit{time tokens} that accurately represent them at their occurrence time.

\begin{figure*}
  \centering
  \includegraphics[width=\textwidth]{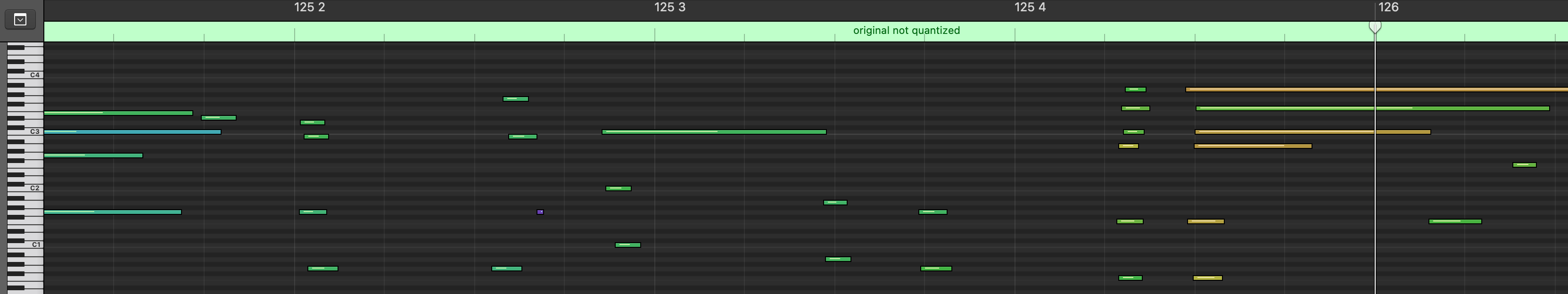}\\
  \vspace{1em}
  \includegraphics[width=\textwidth]{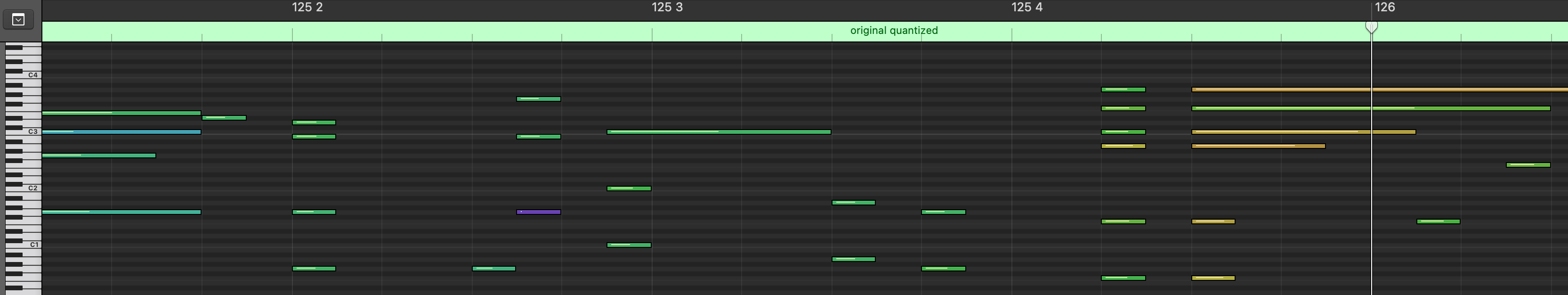}

  \caption[Pianoroll visualization of the MidiTok preprocessing step]{Two pianoroll visualizations of a track of piano: top) original MIDI track, as performed by a human; bottom) the same track preprocessed, with onset and offset times aligned to the 8th of beat.}
  \label{fig:midi_preprocessing}
\end{figure*}

Another important aspect is the precision to which represent this information. Values such as Program (the id of an instrument), pitch, velocity range from 0 to 127. They can be considered as "semi-continuous", as 128 is an arguably large number of possible values. But using all these ranges of possible values with a language model is usually not optimal, as the later is a discrete model and so could struggle to efficiently capture the differences between two consecutive values, e.g. a velocity of 100 from a velocity of 101. It is then usually recommended to "discretize" these values, by downsampling the number of possible values. The set of velocities could for example be downsampled to 16 values, equally spaced between 0 to 127. The same downsampling logic should be applied to the time of all the events: instead of aligning the time with a resolution of e.g. 384 ticks per beat, we quantize the times of the events to a resolution of e.g. 8 ticks per beat, which must be sufficiently precise to keep the overall dynamic of the music. An time downsampling example is depicted in \Cref{fig:midi_preprocessing}. Downsampling these values require a careful preprocessing, before representing them as tokens.

Finally, this information can be represented by different manners. In the MIDI protocol, the moments when the notes are played and when their associated key is released come respectively as the form of \texttt{NoteOn} and \texttt{NoteOff} messages / events. We can then choose to directly tokenize these events, or to explicitly represent the durations of the notes by deducing their values from the difference of time between these messages. The time itself can be represented by using \texttt{TimeShift} tokens indicating time movements, or placing \texttt{Bar} and \texttt{Position} tokens indicating respectively the beginning of the new bar and the current position within the current bar. The way to represent several tracks of different instruments is also left to several "design" choices.
This liberty led researchers to introduce different symbolic music tokenizations, each with their benefits, and us to implement them under a unified API in a flexible and user-friendly library that we called MidiTok.

\subsection{Previous works}

Early works using discrete models for symbolic music, such as DeepBach \cite{deepbach2017hadjeres}, FolkRNN \cite{folkrnn2015sturm} or BachBot \cite{bachBot2017}, rely on specific tokenizations specifically designed for the data being used, for instance for the four voices of Bach chorales. Non-autoregressive models such as MuseGAN \cite{dong2017musegan} often represent music as pianoroll matrices. Since then, researchers introduced more universal tokenizations for any kind of music. These strategies represent note attributes and time in a general way that can be used with any music. The most commonly used are \textit{Midi-Like} \cite{oore_midilike_2018} and \textit{REMI} \cite{huang_remi_2020}. The former tokenizes music by representing tokens as the same types of events from the MIDI protocol, while the latter represents time with \textit{Bar} and \textit{Position} tokens and note durations with explicit \textit{Duration} tokens. Additionally, \textit{REMI} includes tokens with additional information such as chords and tempo.

More recently, researchers have focused on improving the efficiency of models with new tokenizations techniques: \textit{Compound Word} \cite{cpword2021}, \textit{Octuple} \cite{zeng2021musicbert} and \textit{PopMAG} \cite{popmag2020} merge embedding vectors before passing them to the model; 2) LakhNES \cite{lakhnes2019} and \cite{MuseNet}, SymphonyNet \cite{symphonynet} and \cite{bpe-for-symbolique-music2023} use tokens combining several values, such as pitch and vocabulary.

\section{MidiTok workflow}

We introduce in this section the base functioning of MidiTok.


First, any tokenizer has to be created from a \texttt{TokenizerConfig} objects. This configuration holds the parameters defining what type of information will be tokenized, and with which precision. A user can choose wether to tokenize tempos, time signature, rests... or not. He can also decide the resolution of values such velocity, time, or the pitch range to tokenize. From this configuration, the tokenizer will create its vocabulary of tokens.
A tokenizer can be saved as a json file, and loaded back as identical without having to provide a configuration.

We consider three categories of tokens: 1)~Global MIDI tokens, which represent attributes and events affecting the music globally, such as the tempo or time signature; 2)~Track tokens, representing values of distinct tracks such as the notes, chords or effects; 3)~Time tokens, which serve to structure and place the previous categories of tokens in time. The categorization of tokens into these three types is important as it will affect the way the tokenizer represents the time.

When tokenizing MIDI tracks, we distinct two modes: a "one token stream" mode which converts all the tracks under a unique sequence of tokens, and a "one stream per track" mode which converts each track independently. In the former mode, the time tokens are created for all the global and track tokens at once, while in the later they are created for each sequence of track token independently.

The tokenization workflow of a MIDI is as follow:
\begin{enumerate}[nosep,label=\textbf{\arabic*.}]
  \item Preprocesses the MIDI object:
  \begin{itemize}[leftmargin=1em, label=-]
    \setlength\itemsep{0em}
    \item If in "one token stream" mode, merges tracks of the same program / instrument;
    \item Removes notes with pitches outside the tokenizer's range, downsample note velocities, onset and offset times;
    \item Downsamples tempo values and times;
    \item Downsamples time signature times;
    \item Removes duplicated notes, tempo and time signature changes, and empty tracks;
  \end{itemize}
  \item Gathers global MIDI events (tempo...);
  \item Gathers track events (notes, chords);
  \item If "one token stream", concatenates all global and track events and sort them by time of occurrence. Else, concatenates the global events to each sequence of track events;
  \item Deduces the time events for all the sequences of events (only one if "one token stream");
  \item Returns the tokens, as a combination of list of strings and list of integers (token ids).
\end{enumerate}

The first and last step are performed the same way for all tokenizers, while other steps can be performed differently depending on the tokenization.
The preprocessing step is essential as it formats the information of a MIDI to fit to the parameters of the tokenizer. The onset and offset times of the track and global events are aligned to the time resolution of the tokenizer, as well as their values. This assures us to retrieve the exact same preprocessed MIDI when detokenizing a the tokens.

\section{Music tokenizations}

\begin{figure}
  \centering
  \includegraphics[width=\columnwidth]{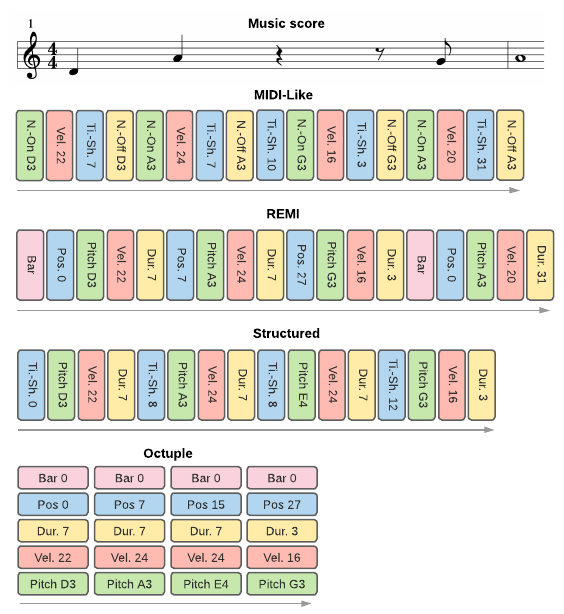}
  \caption{A sheet music and several token representations.}
  \label{fig:miditok_tokens}
\end{figure}

MidiTok implements the most commonly used music tokenizations:

\begin{itemize}[leftmargin=1em]
  \setlength\itemsep{0em}
  \item \textbf{MIDILike} \cite{oore_midilike_2018}: represents MIDI messages as tokens. Notes are represented with \texttt{NoteOn} and \texttt{NoteOff} tokens, indicating their onset and offset times, and time is represented with \texttt{TimeShift} tokens;
  \item \textbf{REMI} \cite{huang_remi_2020}: standing for \textit{Revamped MIDI}, it represents note duration with explicit \texttt{Duration} tokens in place of the \texttt{NoteOff} offset tokens, and time as a combination of \texttt{Bar} and \texttt{Position} tokens indicating respectively the beginning of a new bar and the position of the time within the current bar;
  \item \textbf{REMI+} \cite{vonrutte2022figaro}: is an extension of \textit{REMI}, allowing to also represent note instruments and time signature;
  \item \textbf{Structured} \cite{pia2021hadjeres}: similar to \textit{MIDILike}, except that it represents note durations explicitly as \textit{REMI} and always use the same scheme of token type succession: \texttt{Pitch}, \texttt{Velocity}, \texttt{Duration} and \texttt{TimeShift};
  \item \textbf{TSD} \cite{bpe-for-symbolique-music2023}: standing for \textit{TimeShift \& Duration}, it is identical to \texttt{MIDILike} but with explicit \texttt{Duration} tokens;
  \item \textbf{Compound Word} \cite{cpword2021}: is similar to REMI, but merges several categories of token embeddings in order to reduce the sequence length for the model. For instance, the \texttt{Pitch}, \texttt{Velocity} and \texttt{Duration} embeddings of a note are first concatenated and projected to get a merged embedding, and several output layers are used to predict these several attributes all at once. We qualify such tokenization as \textit{multi-vocabulary}, as in essence it is based on several distinct vocabularies;
  \item \textbf{Octuple} \cite{zeng2021musicbert}: also a \textit{multi-vocabulary} tokenization, it works by merged the embeddings of the attributes of each note, along with the \texttt{Bar\_n} and \texttt{Position\_p} embeddings representing its position in time, resulting in a token sequence as long as the number of notes tokenized;
  \item \textbf{MuMIDI} \cite{popmag2020}: a \textit{multi-vocabulary} tokenization similar to \textit{Compound Word}, but also representing note programs (instruments) along with a built-in positional encoding mechanism based on the number of elapsed bar and position;
  \item \textbf{MMM} \cite{mmm2020}: a tokenization for multitrack music inpainting.
\end{itemize}

\Cref{tab:miditok_decomposing_tokenization} shows a comparative analysis of these tokenizations, and \Cref{fig:miditok_tokens} shows different tokenizations applied on an example music sheet melody.
In these original works, the authors praise the benefits of using what we call \textit{additional tokens}, which represent information other than the notes and time. Also, multi-vocabulary tokenizations were introduced with the main goal to reduce the length of the sequence of embeddings processed by the model, which can be a bottleneck with Transformer models for which the complexity grows quadratically with.
We built MidiTok to offer users the flexibility to choose the types of additional tokens they want to use, and also the possibility to use Byte Pair Encoding for non-multi-vocabulary tokenization to drastically reduce their token sequence lengths.
We introduce these features, and more, in the next section.

\begin{table*}
  \resizebox{\textwidth}{!}{
  \begin{tabular}{lccccccccccccc}
    \toprule
     & \multicolumn{2}{c}{\textbf{Time}} & \multicolumn{2}{c}{\textbf{Note duration}} \\
    \cmidrule(l){2-3} \cmidrule(l){4-5}
    \textbf{Tokenization} & \texttt{TimeShift} & \texttt{Bar} + \texttt{Pos.} & \texttt{Duration} & \texttt{NoteOff} & \textbf{Multitrack} & \textbf{One stream} & \textbf{Multi-voc} & \textbf{Chord} & \textbf{Rest} & \textbf{Tempo} & \textbf{Time Sig.} & \textbf{Pedal} & \textbf{Pit. Bend} \\
    \midrule
    \textbf{MIDI-Like} \cite{oore_midilike_2018} & $\surd$ & - & - & $\surd$ & $\dagger$ & $\dagger$ & - & $\ddagger$ & $\ddagger$ & $\ddagger$ & $\ddagger$ & $\ddagger$ & $\ddagger$ \\
    \textbf{REMI} \cite{huang_remi_2020} & - & $\surd$ & $\surd$ & - & $\dagger$ & $\dagger$ & - & $\ddagger$ & $\ddagger$ & $\ddagger$ & $\ddagger$ & $\ddagger$ & $\ddagger$ \\
    \textbf{Structured} \cite{pia2021hadjeres} & $\surd$ & - & $\surd$ & - & $\dagger$ & $\dagger$ & - & - & - & - & - & - & - \\
    \textbf{TSD} \cite{bpe-for-symbolique-music2023} & $\surd$ & - & $\surd$ & - & $\dagger$ & $\dagger$ & - & $\ddagger$ & $\ddagger$ & $\ddagger$ & $\ddagger$ & $\ddagger$ & $\ddagger$ \\
    \textbf{CP Word} \cite{cpword2021} & - & $\surd$ & $\surd$ & - & $\dagger$ & $\dagger$ & $\surd$ & $\ddagger$ & $\ddagger$ & $\ddagger$ & $\ddagger$ & - & - \\
    \textbf{Octuple} \cite{zeng2021musicbert} & - & $\surd$ & $\surd$ & - & $\surd$ & $\surd$ & $\surd$ & - & - & $\ddagger$ & $\ddagger$ & - & - \\
    \textbf{MuMIDI} \cite{popmag2020} & - & $\surd$ & $\surd$ & - & $\surd$ & $\surd$ & $\surd$ & $\ddagger$ & - & $\ddagger$ & - & - & - \\
    \textbf{MMM} \cite{mmm2020} & $\surd$ & $\surd$ & - & $\surd$ & $\surd$ & $\surd$ & - & $\ddagger$ & - & $\ddagger$ & $\ddagger$ & - & - \\
    \bottomrule
  \end{tabular}}
  \caption[Comparative table of the tokenizations implemented by MidiTok]{Comparative table of the tokenizations implemented by MidiTok. \dag: is true when the tokenizer is configured to represent \texttt{Program} tokens; \ddag: Is optional. \textit{REMI+} \cite{vonrutte2022figaro} is implemented under \textit{REMI}. In MidiTok, MMM is implemented using \texttt{Duration} tokens.}
  \label{tab:miditok_decomposing_tokenization}
\end{table*}

\section{Features}

\subsection{Additional tokens}

MidiTok allows to choose a set of additional tokens to use. These tokens add more musical information, that can be useful in some cases to model:

\begin{itemize}[leftmargin=1em]
  \setlength\itemsep{0em}
  \item Chord: track token tokens describing the chord formed by the following notes. This type of token can help to explicitly model the harmony formed by chords;
  \item Programs: track token informing of the program, or instrument, of the following notes. This token is natively used in some tokenizations, and allows for others to tokenize multiple tracks all at once in a multitrack way;
  \item Sustain pedal: track token representing the sustain pedal events;
  \item Pitch bend: track token representing the pitch bend events;
  \item Tempo: global token informing of the current tempo, indicating the execution speed;
  \item Time signature: global token informing of the time signature. The value of the time signature directly impact the number of beats present in the bars, and the duration of the beats;
  \item Rest: time token acting as time-shifts, occurring only when no notes are being played. This token emphasize silent moments.
\end{itemize}

Huang et al reports that using \texttt{Tempo} and \texttt{Chord} tokens for a generative Transformer yielded generated results of better quality, that were preferred from human evaluators \cite{huang_remi_2020}.

\subsection{Byte Pair Encoding}

Byte Pair Encoding (BPE) \cite{bpe_original_article} is a data compression technique. It converts the most recurrent successive bytes in a corpus into newly created ones.
BPE is nowadays largely used in the NLP field to build the vocabulary, by automatically creating words and sub-words units from the recurrence of their occurrences within a training corpus \cite{sennrich-etal-2016-neural}. In practice BPE is learned until the vocabulary reaches a target size.

MidiTok allows to use BPE for symbolic music in a simple yet powerful way, to create new tokens that can represent whole notes are successions of notes. Similarly to text, the vocabulary is learned from a corpus of MIDI files. Here however, we use the tokenizer's base vocabulary, that is the set of the basic token representing the information introduced in \Cref{sec:miditok_tokenizing_music}, as "bytes". MidiTok rely on Hugging Face's tokenizers library \footnote{\url{https://github.com/huggingface/tokenizers}} for the BPE training and encoding. The library is implemented in Rust and allow very fast computations. To use it for symbolic music, MidiTok associate each base token to a unique byte that the library can recognize.

BPE allows to drastically reduce the sequence length, while taking benefit of the embedding spaces of models such as Transformers, get better results and a faster inference speed \cite{bpe-for-symbolique-music2023}. \Cref{tab:miditok_bpe_seq_len} shows the sequence length reduction offered by BPE. It also shows how BPE increases tokenization and detokenization times. This time increase is however mitigated by the inference speed gains offered by BPE, as the sequence length is drastically reduced.

The method allows to avoid the drawbacks of multi-vocabulary methods, that are to require to implement multiple input and output modules and use a combination of losses that can lead to unstable learning, slower training, and code adaptations of models and training methods.

\begin{table}
  \centering
  \resizebox{\columnwidth}{!}{
  \begin{tabular}{lcccccccc}
      \toprule
      & \multicolumn{2}{c}{\textbf{Voc. size}} & \multicolumn{2}{c}{\textbf{tokens/beat} $(\downarrow)$} & \multicolumn{2}{c}{\textbf{Tok. time}  $(\downarrow)$} & \multicolumn{2}{c}{\textbf{Detok. time} $(\downarrow)$} \\

      \cmidrule(l){2-3} \cmidrule(l){4-5} \cmidrule(l){6-7} \cmidrule(l){8-9}
      \textbf{Strategy} & TSD & REMI & TSD & REMI & TSD & REMI & TSD & REMI \\
      \midrule
      \textbf{No BPE} & 149 & 162 & 18.5 & 19.1 & 0.174 & 0.151 & 0.031 & 0.039 \\
      \textbf{BPE 1k} & 1k & 1k & 9.3 (-49.5\%) & 10.4 (-45.3\%) & 0.187 & 0.163 & 0.053 & 0.063 \\
      \textbf{BPE 5k} & 5k & 5k & 7.0 (-62.2\%) & 8.5 (-55.2\%) & 0.181 & 0.165 & 0.053 & 0.064 \\
      \textbf{BPE 10k} & 10k & 10k & 6.3 (-66.0\%) & 7.7 (-59.7\%) & 0.183 & 0.164 & 0.052 & 0.065 \\
      \textbf{BPE 20k} & 20k & 20k & 5.8 (-68.9\%) & 6.9 (-63.9\%) & 0.184 & 0.163 & 0.052 & 0.063 \\
      \textbf{CP Word} &  & 188 &  & 8.6 (-54.8\%) &  & 0.169 &  & 0.034 \\
      \textbf{Octuple} &  & 241 &  & 5.2 (-72.6\%) &  & 0.118 &  & 0.035 \\

      \bottomrule
  \end{tabular}}
  \caption[Inference speed of models with and without BPE]{Vocabulary size, average tokens per beat ratio, and average tokenization and decoding times in second on the Maestro dataset \cite{maestro-dataset2019}. \textit{CP Word} and \textit{Octuple} are grouped with \textit{REMI} as they represent time similarly with \texttt{Bar} and \texttt{Position} tokens.}
  \vskip -0.15in
  \label{tab:miditok_bpe_seq_len}
\end{table}

\subsection{Hugging Face Hub integration}

The Hugging Face Hub\footnote{\url{https://huggingface.co/}} is a model and dataset sharing platform which is widely used in the AI community. It allows to freely upload, share and download models and datasets, directly in the code in a very convenient way. Its interactions rely on an open-source Python package named \texttt{huggingface\_hub}\footnote{\url{https://github.com/huggingface/huggingface_hub}}. As it works seamlessly in the Hugging Face ecosystem, especially the \texttt{transformers}\footnote{\url{https://github.com/huggingface/transformers}}\cite{wolf-etal-2020-transformers} or Diffusers libraries\footnote{\url{https://github.com/huggingface/diffusers}}, it stood out and became one of the preferred way to openly share and download models.

Now when downloading a Transformer model, one will need to also download its associated tokenizer to be able to “dialog” with it. Likewise, if one wants to share a models, he will need to share its tokenizer too for people to be able to use it. MidiTok allows to push and download tokenizers in similar way to what is done in the Hugging Face Transformers library.

Internally, MidiTok relies on the \texttt{huggingface\_hub.ModelHubMixin} component. It implements the same methods commonly used in the Hugging Face ecosystem: \\\texttt{save\_pretrained}, \texttt{load\_pretrained} and \texttt{push\_to\_hub}. Relying on this component allows to easily interact with the hub with as less code and logic as possible, and so a minimal maintenance cost. MidiTok only deals with the two "bridges" between these methods and how tokenizers are saved and loaded.

We still note that at the moment or writing, there are few symbolic models shared on the internet globally, including the Hugging Face hub. Before releasing the interoperability between MidiTok and the hub, users needed to manually download and upload their tokenizers. This is inconvenient as these operations are usually performed within some code pipelines, which are often executed on remote servers. We hence hope that these feature will encourage people to share their models and use those from the community.

\subsection{Data augmentation}

Data augmentation is a technique to artificially increases the size of a dataset by applying various transformations on to the existing data. These transformations consist in altering one or several attributes of the original data. In the context of images, they can include operations such as rotation, scaling, cropping or color adjustments. This is more tricky in the case of natural language, where the meaning of the sentences can easily diverge following how the text is modified, but some techniques such as paraphrase generation or back translation can fill this purpose.

The purpose of data augmentation is to introduce variability and diversity into the training data without collecting additional real-world data. Data augmentation can be important and increase a model's learning and generalization, as it exposes it to a wider range of variations and patterns present in the data. In turn it can increases its robustness and decrease overfitting.

MidiTok allows to perform data augmentation, on the MIDI level and token level. Transformations can be made by increasing the values of the velocities and durations of notes, or by shifting their pitches by octaves. Data augmentation is highly recommended to train a model, in order to help a model to learn the global and local harmony of music. In large datasets such as the Lakh \cite{lakh_dataset} or Meta \cite{mmd2021} MIDI datasets, MIDI files can have various ranges of velocity, duration values, and pitch. By augmenting the data, thus creating more diversified data samples, a model can effectively learn to focus on the melody, harmony and music features rather than putting too much attention on specific recurrent token successions.

\subsection{PyTorch data loading}

PyTorch is a very popular DL framework, that is widely used in most research and production works. When training a model, the data must be adequately loaded and processed before being fed. For natural language, text data is usually loaded and tokenized on the fly by the CPU, then moved to the GPU on which the model is running. This task is performed by a \texttt{Dataset}, a \texttt{DataLoader} and a data collator.
For symbolic music however, we cannot find a "commonly" used data loading process as the field is not as developed as NLP.

MidiTok offers universal \texttt{Dataset} classes to load symbolic music, and data collator to be used with. \texttt{DatasetTok} will load in memory (RAM) tokens, either by loading json tokens files or tokenizing MIDI files, and split the overall token sequences into the user's desired minimum and maximum length. The sequence splitting is performed as when tokenizing a MIDI, the token sequence are usually relatively long (in order of thousands), whereas model are usually fed sequences up to a limited number of tokens, often in the range of 100 to 1000.
However, loading all tokens in the memory can be intractable with very large datasets or limited memory. That is why the \texttt{DatasetJsonIO} will load token json files on the fly during the training. This dataset class requires to tokenize a MIDI dataset first, and cannot split the sequences loaded. Hence, if a user want to split the sequences, it must do it before with a \texttt{split\_dataset\_to\_subsequences} method that MidiTok also provides.

The MidiTok data collator allows to pad sequences, add \textit{Beginning Of Sequence} and \textit{End Of Sequence} tokens and shift label sequences. Padding can be done on the left side, which can be handy when generating autoregressively from a model, and required with some libraries such as Transformers \cite{wolf-etal-2020-transformers}.

\subsection{Other useful methods}

Finally, MidiTok feature other useful MIDI manipulation and data extraction methods that can be used for other purposes: chord extraction, overlapping notes fix, track merging by program or category of programs, bar count or note deduplication.

\section{Usage insights}

Since its first version in late 2021, MidiTok gain traction as a simple yet flexible way to tokenize music. It is being used by researchers of the Music Information Retrieval (MIR) community for their works, independent developers, students, and now industrial actors. 

\begin{figure}
  \centering
  \includegraphics[width=\columnwidth]{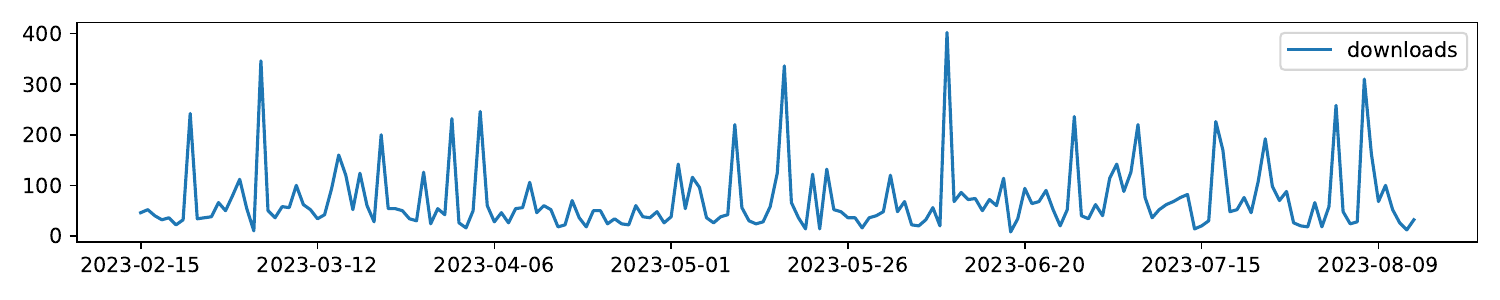}
  \caption[Daily downloads of MidiTok on PyPi]{Daily downloads of MidiTok on PyPi.}
  \label{fig:miditok_downloads}
\end{figure}

As for september 2023, MidiTok gathered more than 450 stars and 50 forks on GitHub. The GitHub repository page counts an average of 100 daily visits, and the package is downloaded on average 900 times per week. \Cref{fig:miditok_downloads} shows the daily downloads of MidiTok on PyPi, for the february 2023 - august 2023 period. It counts a total of 75k downloads since its first release at the time of writing.
We cannot reliably estimate the number of monthly or yearly projects using MidiTok, but can mention that each year a number research papers published at the well recognized ISMIR proceedings are using MidiTok as backbone tokenizer.

Finally, MidiTok is used as backbone in Qosmo's Neutone plugin\footnote{\url{https://neutone.space}}, to be used in DAWs. Neutone allows to use DL models interactively in any DAW as a VST plugin.
This last point may be the most important for the future of MidiTok.
As of 2023, the offer of AI assisted music creation tool has been relatively poor, and most musicians do not use them. We recently witnessed the apparition of text-to-music models such as AudioGen \cite{audiogen} or DiffSound \cite{DiffSound}, capable to generate audio from a text prompt description, but these models are not widely used to generate complete music. While they can produce coherent and high quality results, they face challenges to be used by musicians: their controllability and interactivity are limited, users are not used to create descriptive enough prompts to get the desired results.
Neutone tackles the challenge of interactivity by embedding models right in the production tools of musicians, which we believe is a big step towards a larger adoption of DL models in the creation process of musicians.

\section{Related works}

Other similar projects for symbolic music preprocessing (for deep learning models) have been developed. MusPy \cite{Dong2020MusPy} offers diverse features such as downloading and converting datasets in various formats, analysis and visualization. NoteSeq\footnote{\url{https://github.com/magenta/note-seq}} is also built for music analysis. Both offers methods to tokenize music, but only as \textit{MIDILike} and with limited features.

MidiTok is built with a different vision: to focus solely on tokenization by offering the best features for it, and let the user use other tools better suited for other tasks such as analysis or evaluation.

\section{Conclusion}

Following the growing usage of deep learning models and generative AI, MidiTok stands as an fully-implemented and open source library for symbolic music tokenization. It can easily be used with any other libraries such as Transformers \cite{wolf-etal-2020-transformers}, and is built with powerful and flexible features. It gained traction in the MIR community, and we hope to keep benefiting from its feedback and contributions to further improve the library.



\section*{Ethics Statement}
We believe that open science and open sourcing code and model parameters ensure an equal access to the latest research to everybody. Nevertheless, we acknowledge that generative models can be used in harmful ways to artists and copyright owners.
Generative models can be used to create new content, that can be conditioned on human prompt such as text description. Malevolent users might control them to copy, alter or use content of artist without their approval. Moreover, such model can represent an unfair competitive tool to music creators, which is a time of writing an open issue and subject to ethic considerations.

\section*{Acknowledgements}

We address special thanks to Ilya Borovik and Atsuya Kobayashi for their major contributions, and acknowledge all contributors and people that may have help in any way for the development of MidiTok.

\bibliography{references}

\begin{thebibliography}{30}
\expandafter\ifx\csname natexlab\endcsname\relax\def\natexlab#1{#1}\fi

\bibitem[{Donahue et~al.(2019)Donahue, Mao, Li, Cottrell, and
  McAuley}]{lakhnes2019}
Chris Donahue, Huanru~Henry Mao, Yiting~Ethan Li, Garrison~W. Cottrell, and
  Julian~J. McAuley. 2019.
\newblock \href {http://archives.ismir.net/ismir2019/paper/000083.pdf}
  {Lakhnes: Improving multi-instrumental music generation with cross-domain
  pre-training}.
\newblock In \emph{Proceedings of the 20th International Society for Music
  Information Retrieval Conference, {ISMIR} 2019, Delft, The Netherlands,
  November 4-8, 2019}, pages 685--692.

\bibitem[{Dong et~al.(2020)Dong, Chen, McAuley, and
  Berg-Kirkpatrick}]{Dong2020MusPy}
Hao-Wen Dong, K.~Chen, Julian McAuley, and Taylor Berg-Kirkpatrick. 2020.
\newblock \href {https://api.semanticscholar.org/CorpusID:220969049} {Muspy: A
  toolkit for symbolic music generation}.
\newblock In \emph{International Society for Music Information Retrieval
  Conference}.

\bibitem[{Dong et~al.(2018)Dong, Hsiao, Yang, and Yang}]{dong2017musegan}
Hao-Wen Dong, Wen-Yi Hsiao, Li-Chia Yang, and Yi-Hsuan Yang. 2018.
\newblock \href {https://ojs.aaai.org/index.php/AAAI/article/view/11312}
  {Musegan: Multi-track sequential generative adversarial networks for symbolic
  music generation and accompaniment}.
\newblock \emph{Proceedings of the AAAI Conference on Artificial Intelligence},
  32(1).

\bibitem[{Ens and Pasquier(2020)}]{mmm2020}
Jeff Ens and Philippe Pasquier. 2020.
\newblock \href {http://arxiv.org/abs/2008.06048} {Mmm : Exploring conditional
  multi-track music generation with the transformer}.

\bibitem[{Ens and Pasquier(2021)}]{mmd2021}
Jeffrey Ens and Philippe Pasquier. 2021.
\newblock \href {https://archives.ismir.net/ismir2021/paper/000022.pdf}
  {Building the metamidi dataset: Linking symbolic and audio musical data}.
\newblock In \emph{Proceedings of 22st International Conference on Music
  Information Retrieval, {ISMIR}}.

\bibitem[{Fradet et~al.(2021)Fradet, Briot, Chhel, El~Fallah~Seghrouchni, and
  Gutowski}]{miditok2021}
Nathan Fradet, Jean-Pierre Briot, Fabien Chhel, Amal El~Fallah~Seghrouchni, and
  Nicolas Gutowski. 2021.
\newblock \href {https://github.com/Natooz/MidiTok} {{MidiTok}: A python
  package for {MIDI} file tokenization}.
\newblock In \emph{Extended Abstracts for the Late-Breaking Demo Session of the
  22nd International Society for Music Information Retrieval Conference}.

\bibitem[{Fradet et~al.(2023)Fradet, Gutowski, Chhel, and
  Briot}]{bpe-for-symbolique-music2023}
Nathan Fradet, Nicolas Gutowski, Fabien Chhel, and Jean-Pierre Briot. 2023.
\newblock \href {https://arxiv.org/abs/2301.11975} {{B}yte {P}air {E}ncoding
  for symbolic music}.
\newblock In \emph{Proceedings of the 2023 Conference on Empirical Methods in
  Natural Language Processing}, Singapore. Association for Computational
  Linguistics.

\bibitem[{Gage(1994)}]{bpe_original_article}
Philip Gage. 1994.
\newblock \href {https://dl.acm.org/doi/10.5555/177910.177914} {A new algorithm
  for data compression}.
\newblock \emph{C Users J.}, 12(2):23–38.

\bibitem[{Gardner et~al.(2022)Gardner, Simon, Manilow, Hawthorne, and
  Engel}]{mt3}
Joshua~P Gardner, Ian Simon, Ethan Manilow, Curtis Hawthorne, and Jesse Engel.
  2022.
\newblock \href {https://openreview.net/forum?id=iMSjopcOn0p} {{MT}3:
  Multi-task multitrack music transcription}.
\newblock In \emph{International Conference on Learning Representations}.

\bibitem[{Hadjeres et~al.(2017)Hadjeres, Pachet, and
  Nielsen}]{deepbach2017hadjeres}
Ga{\"e}tan Hadjeres, Fran{\c{c}}ois Pachet, and Frank Nielsen. 2017.
\newblock \href {https://proceedings.mlr.press/v70/hadjeres17a.html}
  {{D}eep{B}ach: a steerable model for {B}ach chorales generation}.
\newblock In \emph{Proceedings of the 34th International Conference on Machine
  Learning}, volume~70 of \emph{Proceedings of Machine Learning Research},
  pages 1362--1371. PMLR.

\bibitem[{Hadjeres and Crestel(2021)}]{pia2021hadjeres}
Gaëtan Hadjeres and Léopold Crestel. 2021.
\newblock \href {http://arxiv.org/abs/2107.05944} {The piano inpainting
  application}.

\bibitem[{Hawthorne et~al.(2021)Hawthorne, Simon, Swavely, Manilow, and
  Engel}]{hawthorne2021sequencetosequence}
Curtis Hawthorne, Ian Simon, Rigel Swavely, Ethan Manilow, and Jesse~H. Engel.
  2021.
\newblock \href {https://archives.ismir.net/ismir2021/paper/000030.pdf}
  {Sequence-to-sequence piano transcription with transformers}.
\newblock In \emph{Proceedings of the 22nd International Society for Music
  Information Retrieval Conference, {ISMIR} 2021, Online, November 7-12, 2021},
  pages 246--253.

\bibitem[{Hawthorne et~al.(2019)Hawthorne, Stasyuk, Roberts, Simon, Huang,
  Dieleman, Elsen, Engel, and Eck}]{maestro-dataset2019}
Curtis Hawthorne, Andriy Stasyuk, Adam Roberts, Ian Simon, Cheng-Zhi~Anna
  Huang, Sander Dieleman, Erich Elsen, Jesse Engel, and Douglas Eck. 2019.
\newblock \href {https://openreview.net/forum?id=r1lYRjC9F7} {Enabling
  factorized piano music modeling and generation with the {MAESTRO} dataset}.
\newblock In \emph{International Conference on Learning Representations}.

\bibitem[{Hsiao et~al.(2021)Hsiao, Liu, Yeh, and Yang}]{cpword2021}
Wen-Yi Hsiao, Jen-Yu Liu, Yin-Cheng Yeh, and Yi-Hsuan Yang. 2021.
\newblock \href {https://doi.org/10.1609/aaai.v35i1.16091} {Compound word
  transformer: Learning to compose full-song music over dynamic directed
  hypergraphs}.
\newblock \emph{Proceedings of the AAAI Conference on Artificial Intelligence},
  35(1):178--186.

\bibitem[{Huang et~al.(2019)Huang, Vaswani, Uszkoreit, Simon, Hawthorne,
  Shazeer, Dai, Hoffman, Dinculescu, and Eck}]{musictransformer2018}
Cheng-Zhi~Anna Huang, Ashish Vaswani, Jakob Uszkoreit, Ian Simon, Curtis
  Hawthorne, Noam Shazeer, Andrew~M. Dai, Matthew~D. Hoffman, Monica
  Dinculescu, and Douglas Eck. 2019.
\newblock \href {https://openreview.net/forum?id=rJe4ShAcF7} {Music
  transformer}.
\newblock In \emph{International Conference on Learning Representations}.

\bibitem[{Huang and Yang(2020)}]{huang_remi_2020}
Yu-Siang Huang and Yi-Hsuan Yang. 2020.
\newblock \href {https://doi.org/10.1145/3394171.3413671} {Pop music
  transformer: Beat-based modeling and generation of expressive pop piano
  compositions}.
\newblock In \emph{Proceedings of the 28th ACM International Conference on
  Multimedia}, MM '20, page 1180–1188, New York, NY, USA. Association for
  Computing Machinery.

\bibitem[{Kreuk et~al.(2023)Kreuk, Synnaeve, Polyak, Singer, D{\'e}fossez,
  Copet, Parikh, Taigman, and Adi}]{audiogen}
Felix Kreuk, Gabriel Synnaeve, Adam Polyak, Uriel Singer, Alexandre
  D{\'e}fossez, Jade Copet, Devi Parikh, Yaniv Taigman, and Yossi Adi. 2023.
\newblock \href {https://openreview.net/forum?id=CYK7RfcOzQ4} {Audiogen:
  Textually guided audio generation}.
\newblock In \emph{The Eleventh International Conference on Learning
  Representations}.

\bibitem[{Liang et~al.(2017)Liang, Gotham, Johnson, and Shotton}]{bachBot2017}
Feynman~T. Liang, Mark Gotham, Matthew Johnson, and Jamie Shotton. 2017.
\newblock \href
  {https://ismir2017.smcnus.org/wp-content/uploads/2017/10/156\_Paper.pdf}
  {Automatic stylistic composition of bach chorales with deep {LSTM}}.
\newblock In \emph{Proceedings of the 18th International Society for Music
  Information Retrieval Conference, {ISMIR} 2017, Suzhou, China, October 23-27,
  2017}, pages 449--456.

\bibitem[{Liu et~al.(2022)Liu, Dong, Cheng, Zhang, Li, Yu, and
  Sun}]{symphonynet}
Jiafeng Liu, Yuanliang Dong, Zehua Cheng, Xinran Zhang, Xiaobing Li, Feng Yu,
  and Maosong Sun. 2022.
\newblock \href {https://arxiv.org/abs/2205.05448} {Symphony generation with
  permutation invariant language model}.
\newblock In \emph{Proceedings of the 23rd International Society for Music
  Information Retrieval Conference}, Bengalore, India. ISMIR.

\bibitem[{Oore et~al.(2018)Oore, Simon, Dieleman, Eck, and
  Simonyan}]{oore_midilike_2018}
Sageev Oore, Ian Simon, Sander Dieleman, Douglas Eck, and Karen Simonyan. 2018.
\newblock \href {https://link.springer.com/article/10.1007/s00521-018-3758-9}
  {This time with feeling: Learning expressive musical performance}.
\newblock \emph{Neural Computing and Applications}, 32:955–967.

\bibitem[{Payne(2019)}]{MuseNet}
Christine Payne. 2019.
\newblock \href {https://openai.com/blog/musenet} {Musenet}.

\bibitem[{Raffel(2016)}]{lakh_dataset}
Colin Raffel. 2016.
\newblock \href {https://colinraffel.com/projects/lmd/} {\emph{Learning-Based
  Methods for Comparing Sequences, with Applications to Audio-to-MIDI Alignment
  and Matching}}.
\newblock Ph.D. thesis, Columbia University.

\bibitem[{Ren et~al.(2020)Ren, He, Tan, Qin, Zhao, and Liu}]{popmag2020}
Yi~Ren, Jinzheng He, Xu~Tan, Tao Qin, Zhou Zhao, and Tie-Yan Liu. 2020.
\newblock \href {https://doi.org/10.1145/3394171.3413721} {Popmag: Pop music
  accompaniment generation}.
\newblock In \emph{Proceedings of the 28th ACM International Conference on
  Multimedia}, page 1198–1206. Association for Computing Machinery.

\bibitem[{Sennrich et~al.(2016)Sennrich, Haddow, and
  Birch}]{sennrich-etal-2016-neural}
Rico Sennrich, Barry Haddow, and Alexandra Birch. 2016.
\newblock \href {https://doi.org/10.18653/v1/P16-1162} {Neural machine
  translation of rare words with subword units}.
\newblock In \emph{Proceedings of the 54th Annual Meeting of the Association
  for Computational Linguistics (Volume 1: Long Papers)}, pages 1715--1725,
  Berlin, Germany. Association for Computational Linguistics.

\bibitem[{Sturm et~al.(2015)Sturm, Santos, and Korshunova}]{folkrnn2015sturm}
Bob~L. Sturm, João~Felipe Santos, and Iryna Korshunova. 2015.
\newblock \href {https://ismir2015.ismir.net/LBD/LBD13.pdf} {Folk music style
  modelling by recurrent neural networks with long short-term memory units}.
\newblock In \emph{Extended abstracts for the Late-Breaking Demo Session of the
  16th International Society for Music Information Retrieval Conference}.

\bibitem[{Vaswani et~al.(2017)Vaswani, Shazeer, Parmar, Uszkoreit, Jones,
  Gomez, Kaiser, and Polosukhin}]{attention_is_all_you_need}
Ashish Vaswani, Noam Shazeer, Niki Parmar, Jakob Uszkoreit, Llion Jones,
  Aidan~N Gomez, \L~ukasz Kaiser, and Illia Polosukhin. 2017.
\newblock \href
  {https://proceedings.neurips.cc/paper/2017/file/3f5ee243547dee91fbd053c1c4a845aa-Paper.pdf}
  {Attention is all you need}.
\newblock In \emph{Advances in Neural Information Processing Systems},
  volume~30. Curran Associates, Inc.

\bibitem[{von R{\"u}tte et~al.(2023)von R{\"u}tte, Biggio, Kilcher, and
  Hofmann}]{vonrutte2022figaro}
Dimitri von R{\"u}tte, Luca Biggio, Yannic Kilcher, and Thomas Hofmann. 2023.
\newblock \href {https://openreview.net/forum?id=NyR8OZFHw6i} {{FIGARO}:
  Controllable music generation using learned and expert features}.
\newblock In \emph{The Eleventh International Conference on Learning
  Representations}.

\bibitem[{Wolf et~al.(2020)Wolf, Debut, Sanh, Chaumond, Delangue, Moi, Cistac,
  Rault, Louf, Funtowicz, Davison, Shleifer, von Platen, Ma, Jernite, Plu, Xu,
  Scao, Gugger, Drame, Lhoest, and Rush}]{wolf-etal-2020-transformers}
Thomas Wolf, Lysandre Debut, Victor Sanh, Julien Chaumond, Clement Delangue,
  Anthony Moi, Pierric Cistac, Tim Rault, Rémi Louf, Morgan Funtowicz, Joe
  Davison, Sam Shleifer, Patrick von Platen, Clara Ma, Yacine Jernite, Julien
  Plu, Canwen Xu, Teven~Le Scao, Sylvain Gugger, Mariama Drame, Quentin Lhoest,
  and Alexander~M. Rush. 2020.
\newblock \href {https://www.aclweb.org/anthology/2020.emnlp-demos.6}
  {Transformers: State-of-the-art natural language processing}.
\newblock In \emph{Proceedings of the 2020 Conference on Empirical Methods in
  Natural Language Processing: System Demonstrations}, pages 38--45, Online.
  Association for Computational Linguistics.

\bibitem[{Yang et~al.(2023)Yang, Yu, Wang, Wang, Weng, Zou, and Yu}]{DiffSound}
Dongchao Yang, Jianwei Yu, Helin Wang, Wen Wang, Chao Weng, Yuexian Zou, and
  Dong Yu. 2023.
\newblock \href {https://doi.org/10.1109/TASLP.2023.3268730} {Diffsound:
  Discrete diffusion model for text-to-sound generation}.
\newblock \emph{IEEE/ACM Transactions on Audio, Speech, and Language
  Processing}, 31:1720--1733.

\bibitem[{Zeng et~al.(2021)Zeng, Tan, Wang, Ju, Qin, and
  Liu}]{zeng2021musicbert}
Mingliang Zeng, Xu~Tan, Rui Wang, Zeqian Ju, Tao Qin, and Tie-Yan Liu. 2021.
\newblock \href {https://doi.org/10.18653/v1/2021.findings-acl.70}
  {{M}usic{BERT}: Symbolic music understanding with large-scale pre-training}.
\newblock In \emph{Findings of the Association for Computational Linguistics:
  ACL-IJCNLP 2021}, pages 791--800, Online. Association for Computational
  Linguistics.

\end{thebibliography}
\bibliographystyle{acl_natbib}

\end{document}